\documentclass[letterpaper, 10 pt, conference]{ieeeconf}

\IEEEoverridecommandlockouts	
\usepackage{cite}
\usepackage{amsmath,amssymb,amsfonts}
\usepackage{algorithmic}
\usepackage{textcomp}
\usepackage{graphicx,color}
\usepackage{mathrsfs}
\usepackage[vlined,ruled]{algorithm2e}
\usepackage{subfigure}
\usepackage{url}
\usepackage{color}
\usepackage{dsfont}
\usepackage{bbm}
\usepackage{booktabs}
\usepackage{array}
\usepackage[table]{xcolor} 
\usepackage{yfonts}
\usepackage[normalem]{ulem}
\usepackage{arydshln}
\usepackage{bbm}

\usepackage[colorlinks]{hyperref}

\newtheorem{theorem}{Theorem}[section]

\newtheorem{definition}{Definition}

\newtheorem{remark}{Remark}

\newtheorem{example}{Example}

\newcommand{\setdef}[2]{\{#1 \; : \; #2\}}

\usepackage{tabularx}
\usepackage{multirow}
\usepackage{makecell}

\newcommand\aamsout{\bgroup\markoverwith{\textcolor{violet}{\rule[0.5ex]{2pt}{1pt}}}\ULon}

\newcommand{\real}{\mathbb{R}}

\newcommand{\transpose}{\mathsf{T}} 

\newcommand{\mc}{\mathcal}

\newcommand{\lip}{\mathrm{lip}}
\newcommand{\Lip}{\mathrm{Lip}}

\newcommand{\1}{\mathds{1} }

\DeclareSymbolFont{bbold}{U}{bbold}{m}{n}
\DeclareSymbolFontAlphabet{\mathbbold}{bbold}

\newcommand{\enom}{e_{\text{nom}}}
\newcommand{\eadv}{e_{\text{adv}}}
\newcommand{\Ntrain}{N_{\text{train}}}
\newcommand{\Ntest}{N_{\text{test}}}

\newcommand\oprocendsymbol{\hbox{$\square$}}
\newcommand\oprocend{\relax\ifmmode\else\unskip\hfill\fi\oprocendsymbol}

\renewcommand{\theenumi}{(\roman{enumi})}

\newcommand*{\QEDA}{\hfill\ensuremath{\blacksquare}}%

\graphicspath{{figs/}}

\renewcommand{\baselinestretch}{0.97432}

\begin{document}

\title{\bf }
\title{\bf\huge\textcolor{black}{Robust Adversarial Classification via Abstaining}}


\author{Abed~AlRahman~Al~Makdah,~Vaibhav~Katewa,~and~Fabio~Pasqualetti%
  \thanks{This material is based upon work supported in part by awards
    ONR ONR-N00014-19-1-2264, AFOSR FA9550-19-1-0235 and
    FA9550-20-1-0140. A. A. Al Makdah and F. Pasqualetti are with the
    Department of Electrical and Computer Engineering and the
    Department of Mechanical Engineering at the University of
    California, Riverside, respectively,
    \href{mailto:aalmakdah@engr.ucr.edu}{\{\texttt{aalmakdah}},\href{mailto:fabiopas@engr.ucr.edu}{\texttt{fabiopas\}@engr.ucr.edu}}.
    V. Katewa is with the Department of Electrical Communication
    Engineering at the Indian Institute of Science, Bangalore, India,
    \href{mailto:vkatewa@iisc.ac.in}{\texttt{vkatewa@iisc.ac.in}}.}}
    
\maketitle

\begin{abstract}
In this work, we consider a binary classification problem and cast it into a binary hypothesis testing framework, where the observations can be perturbed by an adversary. To improve the adversarial robustness of a classifier, we include an abstain option, where the classifier abstains from making a decision when it has low confidence about the prediction. We propose metrics to quantify the nominal performance of a classifier
  with an abstain option and its robustness against adversarial
  perturbations. We show that there exist a tradeoff between the two
  metrics regardless of what method is used to choose the abstain
  region. Our results imply that the robustness of a classifier with an abstain option can
  only be improved at the expense of its nominal
  performance. Further, we provide necessary conditions to design the
  abstain region for a $1$-dimensional binary classification
  problem. We validate our theoretical results on the MNIST dataset,
  where we numerically show that the tradeoff between performance and
  robustness also exist for the general multi-class classification
  problems.
\end{abstract}

%

\section{Introduction}\label{sec: introduction}

Data-driven and machine learning models are shown to be vulnerable to adversarial examples, which are small, targeted, and malicious perturbations of the inputs that induce unwanted, and possibly dangerous model behavior \cite{CS-WZ-IS-JB-DE-IG-RF:14}. For instance, placing stickers at specific locations on a stop sign can fool a state-of-the-art model into classifying it as a speed limit sign \cite{KE-IE-EF-BL-AR-CX-AP-TK-DS:18}. This vulnerability is one of the main limitations that hurdle the deployment of data-driven systems in safety-critical applications, such as medical diagnosis \cite{AE-BK-RAN-JK-SMS-HMB-ST:17}, robotic surgery \cite{AS-RSD-JDO-SL-AK-PCWK:16}, and self-driving cars \cite{MB-DDT-DD-BF-BF-PG-LDJ-MM-UM-JZ-XZ-JZ-KZ:16}.  In control applications, classifiers play an important role in decision making, in particular for autonomous systems \cite{MB-DDT-DD-BF-BF-PG-LDJ-MM-UM-JZ-XZ-JZ-KZ:16,KDJ-JL-JSB-MPO-MJK:16,PZ-JI-BF-SF:17}. Unlike data-driven models that help with writing an email, classify images of cats and dogs, or recommend movies, small error in safety-critical applications can result in catastrophic consequences \cite{SL:16}.  A substantial body of literature addresses adversarial robustness of data-driven models \cite{AM-AM-LS-DT-AV:18,AAALM-VK-FP:19a,VK-AAALM-FP:20,HZ-YY-JJ-EPX-LEG-MIJ:19b}. Despite all these contributions to guarantee robustness against adversarial perturbations, robust models still fail to achieve optimal robustness. In fact, improving the robustness of these models comes at the expense of their nominal performance \cite{AAALM-VK-FP:19,HZ-YY-JJ-EPX-LEG-MIJ:19b,DT-SS-LE-AT-AM:19}. Thus, unwanted behavior will still exist for robust models on nominal inputs, therefore, safety remains at risk.

Several frameworks are developed to improve the adversarial robustness in classification \cite{AM-AM-LS-DT-AV:18,AAALM-VK-FP:19a,VK-AAALM-FP:20,HZ-YY-JJ-EPX-LEG-MIJ:19b}. However, in all these frameworks, robustness of a classifier is mainly improved via tuning the position of its decision boundaries. In this work, we take a different route for addressing adversarial robustness in classification problems. We consider an abstain option, where a classifier with fixed classification boundaries may abstain from giving an output over some region in the input space that the classifier is uncertain about. Mainly the inputs in such a region are the most prone to adversarial attacks. Thus, abstaining over such a region helps the classifier to avoid misclassifying perturbed inputs, and hence improve its adversarial robustness. In particular, under a perturbed input, instead of giving a wrong output (or possibly a correct output with low confidence), the model decides to abstain from giving one. For instance, if a self-driving car detects an object that it is uncertain about (it could be a shadow or maybe sensor measurements are perturbed by an adversary), it could abstain from giving an output that might lead to a car accident, and ask a human to take control. In safety critical applications, abstaining on low confidence output might be better than making a wrong decision.
\begin{figure}[!t]
  \centering
  \includegraphics[width=0.8\columnwidth,trim={0cm 0cm 0cm
    0cm},clip]{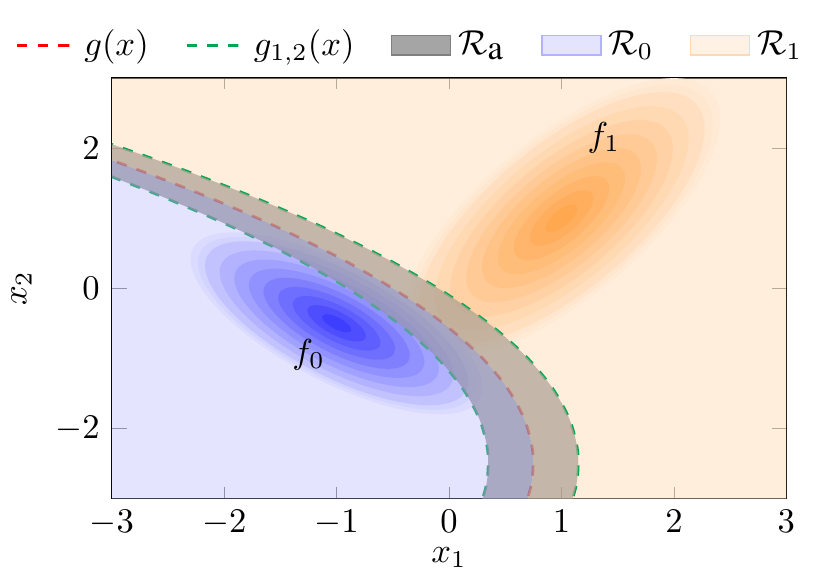}
  \caption{This figure shows the distribution of $x$ under class $\mc{H}_0$ (blue ellipsoid) and class $\mc{H}_1$ (orange ellipsoid). $g(x)$, represented by the dashed red line, is the hyperspace decision boundary for the non-abstain case for the classifier in \eqref{eq: general classifier high dimension}. It divides the observation space into $\mc{R}_0$ (blue region) and $\mc{R}_1$ (orange region). $g_1(x)$ and $g_2(x)$, represented by the green dashed lines, are the boundaries of the abstain region $\mc{R}_{\text{a}}$ (gray region).}
  \label{fig: d_dim_hyp_test}
\end{figure}

Motivated by this, we study the problem of classification with an abstain option by casting it into a binary hypothesis testing framework, where we add a third region in the observation space that corresponds to the observations on which the classifier abstains on (Fig. \ref{fig: d_dim_hyp_test}). Particularly, we study the relation between the accuracy and the adversarial robustness of a binary classifier upon varying the abstain region, where we show that improving the adversarial robustness of a classifier via abstaining comes at the expense of its accuracy.

\textbf{Contributions.}
This paper features three main contributions. First, we propose metrics to quantify the performance of a classifier with an abstain option and its adversarial robustness. Second, we show that for a binary classification problem with an abstain option, a tradeoff between performance and adversarial robustness always exist regardless of which region of the input space is abstained on. Thus, the robustness of a classifier with an abstain option can only be improved at the expense of its nominal performance. Further, we numerically show that such a tradeoff exist for the general multi-class classification problems. The type of the tradeoff we present in this paper is different than the one studied in the literature \cite{AAALM-VK-FP:19,HZ-YY-JJ-EPX-LEG-MIJ:19b,DT-SS-LE-AT-AM:19}, degrading the nominal performance implies that the classifier abstains more often on nominal inputs, and it does not imply an increase in the misclassification rate. Third, we provide necessary conditions to optimally design the abstain region for a given classifier for the $1$-dimensional binary classification problem.
%

\textbf{Related work.} The literature on classification with an abstain option (also referred to as reject option or selective classification) mainly discusses methods on how to abstain on uncertain inputs. \cite{RH-MHW:06,PLB-MHW:08} augmented the output class set with a reject class in a binary classification problem, where inputs with probability below a certain threshold are abstained on. Further, \cite{YG-REY:17} used abstaining in multi-class classification problems, where abstaining was used in deep neural networks. In \cite{AZ-CD-MH:20}, abstaining was used in a regression learning problem. While little work has been done on using abstaining in the context of adversarial robustness, recent work has developed algorithms that guarantee robustness against adversarial attacks via abstaining \cite{CL-SF:19,NB-AB-DS-HZ:21}, where a tradeoff between nominal performance and adversarial robustness has been observed upon tuning their algorithms. In this work, we formally prove the existence of such a tradeoff between performance and adversarial robustness, where we show that this tradeoff exist regardless of what algorithm is used to select the abstain region.
%
%

\textbf{Paper’s organization.} The rest of the paper is organized as follows. Section \ref{sec: setup} contains our mathematical setup. Section \ref{sec:tradeoff} contains the tradeoff between performance and robustness, design of optimal abstain region, and an illustrative example. Section \ref{sec:MNIST} contains our numerical experiment on the MNIST dataset, and Section \ref{sec:conclusion} concludes the paper.

\section{Problem setup and preliminary notions}\label{sec: setup} 
We consider a $d$-dimensional binary classification problem formulated as hypothesis testing problem as in \cite{AAALM-VK-FP:19}. The objective is to decide whether an observation $x \in \real^d$ belongs to class $\mc H_0$ or class $\mc H_1$. We assume that the distribution of the observations under class $\mc{H}_0$ and class $\mc{H}_1$ satisfy 
\begin{align}\label{eq: hypotheses}
  \begin{aligned}
    &\mc H_0: x \sim f_0(x ), \text{ and } \mc H_1: x \sim
    f_1(x),
  \end{aligned} \end{align}
where $f_0(x)$ and $f_1(x)$ are known arbitrary probability density functions. For notational convenience, in the rest of this paper we denote $f_0(x)$ and $f_1(x)$ by $f_0$ and $f_1$, respectively. We denote the prior probabilities of the observations under $f_0$ and $f_1$ by $p_0$ and $p_1$, respectively. In this setup, any classifier can be represented by a partition of the $\real^d$ space by placing decision boundaries at suitable positions (see Fig. \ref{fig: d_dim_hyp_test}).
%
%
%
%
%
We consider adversarial manipulations of the observations, where an attacker is capable of adding perturbations to the observations in order to degrade the performance of the classifier. We model\footnote{In this work, we do not specify a model for the adversary, our analysis holds independently of the adversary model.} such manipulations as a change of the probability density functions in \eqref{eq: hypotheses}. We refer to the perturbed $f_0$ and $f_1$ in \eqref{eq: hypotheses} as $\widetilde{f}_0$ and $\widetilde{f}_1$, respectively. In this work, we aim to improve the adversarial robustness of any classifier by abstaining from making a decision for low confidence outputs. A classifier with an abstain option can be written as
\begin{align}\label{eq: general classifier high dimension}
  \mathfrak{C}(x;g(x),g_1(x),g_2(x)) =
  \begin{cases}
    \mc H_0, & x \in \mc R_0 \cap \overline{\mc R}_{\text{a}},\\
    \mc H_{1}, & x \in \mc R_1 \cap \overline{\mc R}_{\text{a}},\\
    \mc H_{\text{a}}, & x \in \mc R_{\text{a}},
  \end{cases}
\end{align}
where $g(x)$\footnote{Technically, $g(x)$ is not the boundary, $g(x)=0$ provides the boundary, but for the notational convenience we use $g(x)$ to refer to the boundary. Similarly, we use $g_1(x)$ and $g_2(x)$ instead of $g_1(x)=0$ and $g_2(x)=0$.} gives the hyperspace decision boundary for the non-abstain case, $g_1(x)$ and $g_2(x)$ give the hyperspace boundaries for the abstain region, specifically,
\begin{align}\label{eq:regions}
\begin{split}
  \mc R_0 &= \setdef{z}{ g(z)\leq0,  \forall z\in \mathbb{R}^d} ,\\
  \mc R_1 &= \setdef{z}{ g(z) >0, \forall z\in \mathbb{R}^d} ,\\
  \mc R_{\text{a}} &= \setdef{z}{(g_1(z)\geq 0) \cap (g_2(z)\leq 0), \forall z\in \mathbb{R}^d},
  \end{split}
\end{align}
and $\overline{\mc R}_{\text{a}}$ is the complement set of $\mc R_{\text{a}}$. We define two metrics to measure the performance and robustness of classifier \eqref{eq: general classifier high dimension}.
\begin{definition}{\bf \emph{(Nominal error)}}\label{def: nom_error}
The nominal error of a classifier with an abstain option is the proportion of the (unperturbed) observations that are misclassified or abstained~on,
\begin{align}\label{eq:nom_error}
e_{\text{nom}}(\mc{R}_0,\mc{R}_1,\mc{R}_{\text{a}} )=&p_0\mathbf{P} \left[ x \in \mc R_1 | \mc H_0 \right] + p_{1} \mathbf{P} \left[ x \in \mc R_0 | \mc H_{1} \right]\nonumber\\
      &+p_0 \mathbf{P} \left[ x \in (\mc R_0 \cap \mc R_{\text{a} }) | \mc H_0\right]\nonumber\\
      &+ p_{1} \mathbf{P} \left[ x \in (\mc R_1\cap \mc R_{\text{a}}) | \mc H_{1} \right],
\end{align}
where $\mc{R}_0$, $\mc{R}_1$, and $\mc{R}_{\text{a}}$ are as in \eqref{eq:regions}.\oprocend
\end{definition}
The first two terms in \eqref{eq:nom_error} correspond to the error without abstaining, therefore, they do not depend on the abstain region $\mc{R}_{\text{a}}$. The last two terms correspond to the abstain error, thus, they depend on $\mc{R}_{\text{a}}$. Using Definition \ref{def: nom_error} and the distributions in \eqref{eq: hypotheses}, the nominal error for classifier \eqref{eq: general classifier high dimension}~is~written~as
  \begin{align}\label{eq: nom_error_high_dim}
      e_{\text{nom}}(\mc{R}_0,\mc{R}_1,\mc{R}_{\text{a}}  ) &= p_0\int\limits_{\mc R_1} f_0dx+ p_1\int\limits_{\mc R_0} f_1dx \nonumber\\
      &+p_0\int\limits_{\mc R_0\cap \mc R_{\text{a}}} f_0dx + p_1\int\limits_{\mc R_1\cap \mc R_{\text{a}}} f_1dx.
  \end{align}
As can be seen in \eqref{eq: nom_error_high_dim}, the nominal classification error depends on $\mc{R}_0$, $\mc{R}_1$, and $\mc{R}_{\text{a}}$, and thus on the position of the boundaries, $g(x)$, $g_1(x)$, and $g_2(x)$, as described in \eqref{eq:regions}. Lower nominal error implies higher classification performance. Note that, if there is no abstain option ($\mc{R}_{\text{a}}=\varnothing$), then the nominal error is equal to the error computed in the classic hypothesis testing framework \cite{TAS-AAG:06}.
\begin{definition}{\bf \emph{(Adversarial error)}}\label{def: adv_error}
  The adversarial error of a classifier with an abstain option is the proportion of the perturbed observations that are misclassified and not abstained on,
  \begin{align}\label{eq: adv_error}
      e_{\text{adv}}(\mc{R}_0,\mc{R}_1,\mc{R}_{\text{a}}) = &p_0 \mathbf{P} \left[ \widetilde{x} \in (\mc R_1 \cap \overline{\mc R}_{\text{a}})| \mc H_0 \right] \nonumber\\
       &+ p_{1} \mathbf{P} \left[\widetilde{x} \in (\mc R_0 \cap \overline{\mc R}_{\text{a}})| \mc H_{1} \right],
  \end{align}
  where $\widetilde{x}\in\mathbb{R}^d$ is a perturbed observation that follows distributions $\widetilde{f}_0$ and $\widetilde{f}_1$ under classes $\mc{H}_0$ and $\mc{H}_1$, respectively.~\oprocend
   \end{definition}%
Using Definition \ref{def: adv_error} and the distributions in \eqref{eq: hypotheses}, we can write the adversarial error for classifier \eqref{eq: general classifier high dimension} as
  \begin{align}\label{eq: adv_error_high_dim}
      e_{\text{adv}}(\mc{R}_0,\mc{R}_1,\mc{R}_{\text{a}}) =p_0\int\limits_{\mc R_1 \cap \overline{\mc R}_{\text{a}}} \tilde{f}_0dx+p_1\int\limits_{\mc R_0 \cap \overline{\mc R}_{\text{a}}} \tilde{f}_1dx,
  \end{align}
Similar to the nominal error, the adversarial error depends on $\mc{R}_0$, $\mc{R}_1$, and $\mc{R}_{\text{a}}$ defined in \eqref{eq:regions}. Further, the adversarial error depends on the perturbed distributions $\tilde{f}_0$ and $\tilde{f}_1$. The adversarial error is related to the classifier's robustness to adversarial attacks, where low adversarial error implies higher robustness. Note that, if a classifier abstains over the whole input space ($\mc{R}_{\text{a}}=\mathbb{R}^d$), then the adversarial error converges to zero, and the classifier achieves maximum possible robustness. Yet, such classifier achieves maximum~nominal~error.
\begin{remark}{\bf\emph{(Intuition behind Definition \ref{def: nom_error} and \ref{def: adv_error})}}\label{rmrk:intuition}
%
%
Abstaining from making a decision can be better than making a wrong one, yet worse than making a correct one. $e_{\text{nom}}$ penalizes abstaining (along with misclassification) since the classifier is not performing the required task, which is to make a decision. On the other hand, $e_{\text{adv}}$ does not penalize abstaining since by abstaining from making a decision on perturbed inputs, the classifier is avoiding an adversarial attack that can lead to misclassification. Each of these two definitions is a different performance metric, where $e_{\text{nom}}$ measures the classifier's nominal performance, while $e_{\text{adv}}$ measures the classifier's robustness against adversarial perturbations of the input. Further, these definitions guarantee that abstaining does not yield a unilateral advantage or disadvantage, where the classifier would abstain always or never. We remark that different definitions are possible.~\oprocend
\end{remark}
\section{Tradeoff between nominal and adversarial errors}\label{sec:tradeoff}
\begin{figure}[!t]
  \centering
  \includegraphics[width=0.8\columnwidth,trim={0cm 0cm 0cm
    0.08cm},clip]{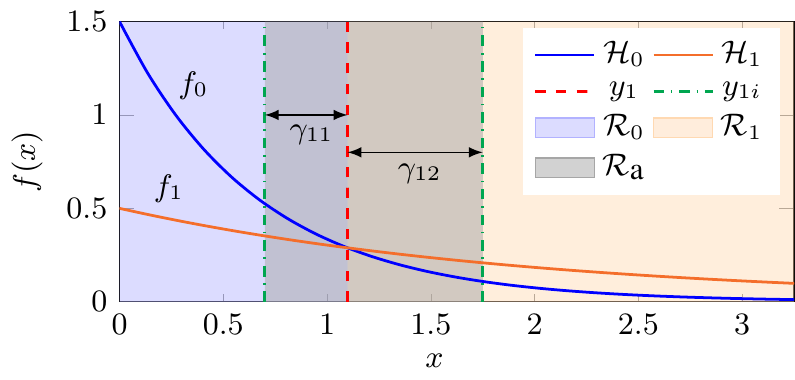}
  \caption{This figure shows the binary classification problem described in Example \ref{ex:exp_dist}, where the observation $x$ under hypotheses $\mc{H}_0$ (solid blue line) and $\mc{H}_1$ (solid orange line) follows exponential distribution~with~$\left. \rho_0=1.5\right.$ and $\rho_1=0.5$, respectively. The dashed red line is the decision boundary for the non-abstain case, which divides the space into $\mc{R}_0$~(blue~region) and $\mc{R}_1$ (orange region). The dot-dashed green lines are the boundaries of the abstain region $\mc{R}_{\text{a}}$ (gray region), which is parametrized by $\gamma_{11}$ and $\gamma_{12}$.} 
  \label{fig:1_dim_hyp_test}
\end{figure}

Ideally, we would like both the nominal error and the adversarial error to be small. However, in this section we show that these errors cannot be minimized simultaneously.
%
%
%
\begin{theorem}{\bf\emph{(Nominal-adversarial error tradeoff)}}\label{thrm:tradeoff}
For classifier \eqref{eq: general classifier high dimension}, let $\mc R_{\text{a}0}=\mc R_0 \cap \mc R_{\text{a}}$ and $\mc R_{\text{a}1}=\mc R_1 \cap \mc R_{\text{a}}$, and let $\widetilde{\mc{R}}_{\text{a}} \supset \mc{R}_{\text{a}}$ be another abstain region that is partitioned as $\widetilde{\mc{R}}_{\text{a}}=\widetilde{\mc{R}}_{\text{a0}} \cup \widetilde{\mc{R}}_{\text{a1}}$, with $\widetilde{\mc R}_{\text{a}0}\supset \mc R_{\text{a}0}$ and $\widetilde{\mc R}_{\text{a}1}\supset \mc R_{\text{a}1}$. Then,
%
%
\begin{align*}
\enom(\mc{R}_0,\mc{R}_1,\mc{R}_{\text{a}})&<\enom(\mc{R}_0,\mc{R}_1,\widetilde{\mc{R}}_{\text{a}}),\\
\eadv(\mc{R}_0,\mc{R}_1,\mc{R}_{\text{a}})&>\eadv(\mc{R}_0,\mc{R}_1,\widetilde{\mc{R}}_{\text{a}}).
\end{align*}
%
%

\begin{proof}
For notational convenience, we denote $e_{\text{nom}}(\mc{R}_{0},\mc{R}_{1},\mc R_{\text{a}} )$, $e_{\text{adv}}(\mc{R}_{0},\mc{R}_{1},\mc R_{\text{a}})$, $e_{\text{nom}}(\mc{R}_{0},\mc{R}_{1},\widetilde{\mc R}_{\text{a}} )$, and $e_{\text{adv}}(\mc{R}_{0},\mc{R}_{1},\widetilde{\mc R}_{\text{a}})$ by $e_{\text{nom}}$, $e_{\text{adv}}$, $\widetilde{e}_{\text{nom}}$, and $\widetilde{e}_{\text{adv}}$, respectively. For a classifier as in \eqref{eq: general classifier high dimension} with abstain region $\widetilde{\mc{R}}_{\text{a}}$, we can write
%
%
%
%
\begin{align*}
\widetilde{e}_{\text{nom}}=&p_0\Big(\int\limits_{\mc R_1} f_0dx+\int\limits_{\widetilde{\mc R}_{\text{a}0}} f_0dx\Big)+ p_1\Big(\int\limits_{\widetilde{\mc R}_{\text{a}1}}f_1dx+\int\limits_{\mc R_0} f_1dx\Big)\\
      =& p_0\int\limits_{\mc R_1} f_0dx+p_0\int\limits_{\mc R_{\text{a}0}} f_0dx+p_0\int\limits_{\widetilde{\mc R}_{\text{a}0} \backslash  \mc R_{\text{a}0}} f_0dx\\
      &+ p_1\int\limits_{\mc R_0} f_1dx+ p_1\int\limits_{\mc R_{\text{a}1}}f_1dx+ p_1\int\limits_{\widetilde{\mc R}_{\text{a}1}\backslash \mc R_{\text{a}1}}f_1dx.
\end{align*}
Then, we can write
\begin{align*}
\widetilde{e}_{\text{nom}}-e_{\text{nom}}=p_0\int\limits_{\widetilde{\mc R}_{\text{a}0} \backslash  \mc R_{\text{a}0}} f_0dx+ p_1\int\limits_{\widetilde{\mc R}_{\text{a}1}\backslash \mc R_{\text{a}1}}f_1dx>0.
\end{align*}
Similarly, we can write
\begin{align*}
\widetilde{e}_{\text{adv}} - e_{\text{adv}}=-p_0\int\limits_{\widetilde{\mc R}_{\text{a}0} \backslash  \mc R_{\text{a}0}} \tilde{f}_0 dx-p_1\int\limits_{\widetilde{\mc R}_{\text{a}1}\backslash \mc R_{\text{a}1}}\tilde{f}_1dx<0.
\end{align*}\end{proof}\end{theorem}
As we increase the abstain region from $\mc R_{\text{a}}$ to $\widetilde{\mc R}_{\text{a}}$, $\enom$ strictly increases, while $\eadv$ strictly decreases, which indicates a tradeoff relation between both errors as we vary the abstain region. Theorem \ref{thrm:tradeoff} implies that there exist a tradeoff between $\enom$ and $\eadv$. Therefore, the classifier's adversarial robustness can be improved only at the expense of its classification performance. In practice, the classifier's robustness can be improved by increasing $\mc{R}_{\text{a}}$, while the nominal classification performance can be improved~by~\mbox{decreasing $\mc{R}_{\text{a}}$}.
%
%
\begin{remark}{\bf\emph{(Comparing our tradeoff with the literature)}}\label{rmrk:tradeoff} \cite{AAALM-VK-FP:19,HZ-YY-JJ-EPX-LEG-MIJ:19b,DT-SS-LE-AT-AM:19} showed that a tradeoff relation exists between a classifier's nominal performance and its adversarial robustness. Despite using different frameworks, their performance-robustness tradeoff relation is obtained via tuning the classifier's boundaries in a way that improves its robustness. In our result, we fix the classifier's decision boundaries, and include an abstain region that can be tuned to obtain our performance-robustness tradeoff. It is possible that a~classifier with an abstain option and a classifier without an abstain option but with different decision boundaries achieve the same $\enom$ and $\eadv$. Although both classifiers achieve same metrics, they are different, where the latter gives an output all the time, while the former abstains~on~some~\mbox{inputs.}~\oprocend
%
\end{remark}
Next we provide our analysis on how to select the abstain region for the $1$-dimensional binary classification problem. Consider the same binary hypothesis testing problem introduced in Section \ref{sec: setup}, but with a scalar observation space where the observation $x\in \mathbb{R}$ is distributed under classes $\mc H_0$ and $\mc H_1$ as in \eqref{eq: hypotheses}. In this setup, any classifier can be represented by a partition of the real line by placing decision boundaries at suitable positions (see Fig. \ref{fig:1_dim_hyp_test}).
%
%
%
%
%
Let\footnote{For simplicity and without loss of generality, we assume that $n$ is~even. Further, an alternative configuration of the classifier \eqref{eq: general classifier high dimension} assigns $\mc{H}_0$ and $\mc{H}_1$ to $\mc{R}_1$ and $\mc{R}_0$, respectively. However, we consider only the configuration in \eqref{eq: general classifier high dimension} without affecting the generality of our analysis.} 
\mbox{$-\infty=y_0 \le \dots \le y_{n+1}= \infty$} denote $n$ decision boundaries with $y = [y_i]$. Then, the classifier regions are
\begin{align*}
  \mc R_0 &= \setdef{z}{ y_i < z < y_{i+1},  \text{ for } i = 0,2,\dots, n} ,\\
  \mc R_1 &= \setdef{z}{ y_i \le z \le y_{i+1}, \text{ for } i = 1,3,\dots, n-1} ,\\
  \mc R_{\text{a}} &= \setdef{z}{ y_{i}-\gamma_{i1} \le z \le y_{i}+\gamma_{i2}, \text{ for } i = 1,2,\dots, n},
\end{align*}
where $\gamma_{ij} \in \real_{\geq 0}$ for $i=1,2,\dots,n$ and $j=1,2$. Let $\left. \gamma=[\gamma_{11},\gamma_{12},\dots,\gamma_{i1},\gamma_{i2},\dots,\gamma_{n1},\gamma_{n2}]^{\transpose}\right.$, $y_{i1}=y_{i}-\gamma_{i1}$, and $y_{i2}=y_{i}+\gamma_{i2}$. Using \eqref{eq:nom_error} and \eqref{eq: hypotheses}, we have
%
\begin{align}\label{eq: nom_error_formula}
e_{\text{nom}}(y,\gamma)=&p_0\left(\sum_{l=1}^{n}(-1)^{l}\int\limits_{-\infty}^{y_{lj}} f_0dx\right)\nonumber\\
&+p_1\left(\sum_{l=1}^{n}(-1)^{l+1}\int\limits_{-\infty}^{y_{lk}} f_1dx+1\right).
\end{align}
where $j=\frac{(-1)^l+1}{2}+1$ and $k=\frac{(-1)^{l+1}+1}{2}+1$ for $l=1,\dots,n$. Using \eqref{eq: adv_error}, the adversarial error becomes
\begin{align}\label{eq: adv_error_formula}
e_{\text{adv}}(y,\gamma)=&p_0\left(\sum_{l=1}^{n}(-1)^{l}\int\limits_{-\infty}^{y_{lk}}\tilde{f}_0dx\right)\nonumber\\
&+p_1\left(\sum_{l=1}^{n}(-1)^{l+1}\int\limits_{-\infty}^{y_{lj}}\tilde{f}_1dx +1\right),
\end{align}
where $j$ and $k$ are the same as above. Given a classifier as in \eqref{eq: general classifier high dimension} with known boundaries $y$, we are interested in how to select the abstain region, i.e., how to choose $\gamma$ given $y$. To this aim, we cast the following optimization problem:
\begin{equation}\label{eq:Opt_prob}
\begin{aligned}
\eadv^*(\zeta)= & \;\underset{\gamma}{\min} \quad e_{\text{adv}}(y,\gamma)\\
&\;~\text{s.t.}  ~~~~e_{\text{nom}}(y,\gamma) \leq \zeta,
\end{aligned}
\end{equation}
where $\zeta \in [\enom(y,0 ),1]$. In what follows, we characterize the solution $\gamma^*$ to \eqref{eq:Opt_prob}. We begin by writing the derivative of the errors in \eqref{eq: nom_error_formula} and \eqref{eq: adv_error_formula} with respect to $\gamma$:
\begin{align}\label{eq:error_derivatives}
\frac{\partial e_{\text{nom}}}{\partial \gamma_{i1}}&=p_q f_q(y_i-\gamma_{i1}),~\;\;\frac{\partial e_{\text{nom}}}{\partial \gamma_{i2}}=p_r f_r(y_i+\gamma_{i2}), \\
\frac{\partial e_{\text{adv}}}{\partial \gamma_{i1}}&=-p_r \tilde{f}_r(y_i-\gamma_{i1}),~\frac{\partial e_{\text{adv}}}{\partial \gamma_{i2}}=-p_q \tilde{f}_q(y_i+\gamma_{i2}),\nonumber
\end{align}
where $q=\frac{(-1)^i+1}{2}$ and $r=\frac{(-1)^{i+1}+1}{2}$ for $i=1,\dots n$. Note that the derivative of $\enom$ with respect to $\gamma$ is strictly positive, while that of $\eadv$ is strictly negative. Thus, $\enom$ increases while $\eadv$ decreases as $\gamma$ increases (i.e., as $\mc{R}_{\text{a}}$ increases), which agrees with the result of Theorem \ref{thrm:tradeoff}. Problem \eqref{eq:Opt_prob} is not convex and it might not exhibit a unique solution. The following theorem characterizes a solution $\gamma^*$ to \eqref{eq:Opt_prob}.
%
%
\begin{theorem}{\bf\emph{(Characterizing the solution to the minimization problem \eqref{eq:Opt_prob})}}\label{thrm:Opt_sol}
Given classifier \eqref{eq: general classifier high dimension} with $1$-dimensional input and known $n$ boundaries $y$, the solution $\gamma^*$ to problem \eqref{eq:Opt_prob} satisfies the following necessary conditions
%
\begin{align}
e_{\text{nom}}(y,\gamma )&=\zeta,\label{eq:Opt_sol_constraint}\\
\frac{\partial e_{\text{adv}}(y,\gamma )}{\partial \gamma_{iu}}.\frac{\partial e_{\text{nom}}(y,\gamma )}{\partial \gamma_{jv}}&=\frac{\partial e_{\text{adv}}(y,\gamma )}{\partial \gamma_{jv}}.\frac{\partial e_{\text{nom}}(y,\gamma )}{\partial \gamma_{iu}},\label{eq:Opt_sol_stationarity}
\end{align}
for $i,j=1,\dots,n$, $i\ne j$, and $u,v=1,2$, where the derivatives of $e_{\text{nom}}$ and $e_{\text{adv}}$ with respect to $\gamma$ are as in \eqref{eq:error_derivatives}.

\begin{proof}
Defining the Lagrange function of \eqref{eq:Opt_prob}
\begin{align}\label{eq:lagrange_function}
\mc{L}(\gamma,\lambda)=e_{\text{adv}}(y,\gamma )+\lambda(e_{\text{nom}}(y,\gamma )-\zeta),
\end{align}
where $\lambda$ is the Karush-Kuhn-Tucker (KKT) multiplier. For notational convenience, we denote $e_{\text{adv}}(y,\gamma)$ and $ e_{\text{nom}}(y,\gamma )$ by $e_{\text{adv}}$ and $e_{\text{nom}}$, respectively. The stationarity KKT condition implies $\frac{\partial}{\partial \gamma}\mc{L}(\gamma,\lambda)=0$, which is written as
\begin{align}\label{eq:stationarity_condition}
\frac{\partial e_{\text{adv}}}{\partial \gamma}=-\lambda \frac{\partial e_{\text{nom}}}{\partial \gamma}.
\end{align}
Using \eqref{eq:stationarity_condition} we write
\begin{align}
-\lambda=\frac{\partial e_{\text{adv}}}{\partial \gamma_{iu}} \Big/ \frac{\partial e_{\text{nom}}}{\partial \gamma_{iu}}=\frac{\partial e_{\text{adv}}}{\partial \gamma_{jv}} \Big/ \frac{\partial e_{\text{nom}}}{\partial \gamma_{jv}},
\end{align}
for $i,j=1,\dots,n$, $i\ne j$, and $u,v=1,2$, which gives us \eqref{eq:Opt_sol_stationarity}. The KKT condition for dual feasibility implies that $\lambda\geq 0$. However, since we have $\frac{\partial e_{\text{adv}}}{\partial \gamma}\ne 0$ and $\frac{\partial e_{\text{nom}}}{\partial \gamma}\ne 0$ from \eqref{eq:error_derivatives}, we get from \eqref{eq:stationarity_condition} that $\lambda>0$. Further, the KKT condition for complementary slackness implies
 $\lambda(e_{\text{nom}}-\zeta)=0$. Since $\lambda>0$, then $e_{\text{nom}}-\zeta=0$, which gives us \eqref{eq:Opt_sol_constraint}.\end{proof}
\end{theorem}
\begin{remark}{\bf\emph{(Location of the abstain region in the observation space)}}\label{rmrk:abs_region}
The abstain region in Theorem \ref{thrm:tradeoff} can be located anywhere in the observation space. However, in Theorem \ref{thrm:Opt_sol}, we assume that the abstain region is located around the decision boundaries. This assumption is fair since the observations near the classifier's boundaries tend to have low classification confidence and are prone to misclassification.~\oprocend
\end{remark}
We conclude this section with an illustrative example.
\begin{example}{\bf\emph{(Classifier with an abstain option for exponential distributions)}}\label{ex:exp_dist}
Consider a 1-D binary hypothesis testing problem, where the observation $x\in\mathbb{R}$ under classes $\mc{H}_0$ and $\mc{H}_1$ follows exponential distributions, i.e., the probability density functions in \eqref{eq: hypotheses} have the form $f_i(x)=\rho_i\exp(-\rho_i x)$ over the domain $x\in \mathbb{R}_{\geq 0}$ with parameter $\rho_i>0$ for $i=0,1$. We consider a single boundary classifier with an abstain option as in \eqref{eq: general classifier high dimension}, with boundary $y_1$ and abstain parameters $\gamma_{11}$ and $\gamma_{12}$ (see Fig. \ref{fig:1_dim_hyp_test}). For simplicity, we model the adversarial manipulations of the observations as perturbation added to the distributions' parameters. We refer to the perturbed parameters as $\tilde{\rho}_0$ and $\tilde{\rho}_1$. Using Theorem~\ref{thrm:Opt_sol}:
\begin{align}\label{eq:opt_cond}
&p_0\exp(-\rho_0( y_1-\gamma_{11}))-p_1\exp(-\rho_1( y_1+\gamma_{12}))+p_1=\zeta, \nonumber\\
&p_1^2 \tilde{\rho}_1 \rho_1 \exp(-\tilde{\rho}_1(y_1-\gamma_{11})-\rho_1(y_1+\gamma_{12}))\nonumber\\
&~~~~=p_0^2 \tilde{\rho}_0 \rho_0 \exp(-\tilde{\rho}_0(y_1+\gamma_{12})-\rho_0(y_1-\gamma_{11})).
\end{align}
For a given classifier with known boundary, $y_1$, and with desired nominal performance, $\zeta$, along with the knowledge of the perturbed distribution parameters $\tilde{\rho}_0$ and $\tilde{\rho}_1$, we can choose the optimal abstain region by solving \eqref{eq:opt_cond} for $\gamma_{11}$ and $\gamma_{12}$. A solution of \eqref{eq:opt_cond} corresponds to a local minima of \eqref{eq:Opt_prob}. Note that the constraint \eqref{eq:Opt_prob} is active (see Theorem \ref{thrm:Opt_sol}), hence we have $\enom(y_1,\gamma^* )=\zeta$. Fig. \ref{fig:tradeoff_exp_dist} shows the values of $\eadv^*$ obtained by solving \eqref{eq:opt_cond} for $\gamma_{11}^*$ and $\gamma_{12}^*$ over the range $\zeta\in[\enom(y_1,0 ),1]$ with $\rho_0=1.5$, $\rho_1=0.5$, $\tilde{\rho}_0=1.2$, $\tilde{\rho}_1=0.7$, and $p_0=p_1=0.5$. Moreover, Fig. \ref{fig:tradeoff_exp_dist} shows the values of $\eadv$ as a function of $\enom$ as $\gamma_{11}$ and $\gamma_{12}$ are varied arbitrarily. Both curves show a tradeoff between $\enom$ and $\eadv$ as predicted by Theorem \ref{thrm:tradeoff}. Further, at each value of $\enom\in (\enom(y_1,0 ),1)$, we observe that \mbox{$\eadv^*(\zeta)<\eadv(y_1,\gamma )$}.
\oprocend
\end{example}
\begin{figure}[!t]
  \centering
  \includegraphics[width=0.7\columnwidth,trim={0cm 0cm 0cm
    0cm},clip]{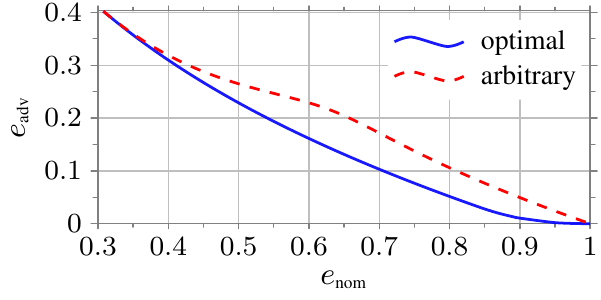}
  \caption{This figure shows the tradeoff between $\enom$ and $\eadv$ as we vary the abstain region, $\gamma$, for the classifier described in Example \ref{ex:exp_dist}. The solid blue line is obtained using Theorem \ref{thrm:Opt_sol} to solve for $\gamma^*$ for each value of $\enom$, while the dashed red line is obtained by varying $\gamma$ arbitrarily. Both curves coincide at the extreme points at $\enom=0.31$ and $\enom=1$, which correspond to $\mc{R}_{\text{a}}=\varnothing$ (no abstaining) and $\mc{R}_{\text{a}}=\mathbb{R}$ (always abstaining), respectively. For $\enom \in (0.31,1)$, we observe that the optimal curve achieves lower $\eadv$ than the curve obtained by arbitrary selection~of~$\gamma$.}
  \label{fig:tradeoff_exp_dist}
\end{figure}
%
%
%
%
%
%
%
%
%
%
\section{Numerical experiment using MNIST dataset}\label{sec:MNIST}
In this section, we illustrate the implications of Theorem \ref{thrm:tradeoff} using the classification of hand-written digits from the MNIST dataset \cite{YL-CC-CJCB:98}. First, we design and train a classifier with an abstain option. Then, we use Definition \ref{def: nom_error} and \ref{def: adv_error} to compute $\enom$ and $\eadv$ for a classifier given the dataset. Finally, we present our numerical results on the MNIST dataset. Although our theoretical results are for binary classification, we show that a tradeoff between $\enom$ and $\eadv$ exists for multi-class classification using the MNIST dataset.
%
\subsection{Classifier design and training}
We design a classifier $h: \mathbb{X} \rightarrow \mathbb{Y}$ using the Lipschitz-constrained loss minimization scheme introduced in \cite{VK-AAALM-FP:20}\footnote{Other classification algorithms, e.g. neural networks, can also be used.}:
\begin{equation}\label{eq:Lip_const_min}
  \begin{aligned}
     \;&\underset{h \in \Lip(\mathbb{X}; \mathbb{Y})}{\min} 
     					\quad \frac{1}{\Ntrain}\sum_{i=1}^{\Ntrain} L \left(h \left( x_i \right), y_i \right), \\
  &~~~\;\text{s.t.} \qquad ~~\lip(h) \leq \alpha,
  \end{aligned}
\end{equation}
where $\mathbb{X}\subset \mathbb{R}^d$ and $\mathbb{Y}\subset\mathbb{R}^m$ are the respective input and output space, $\text{Lip}(\mathbb{X}; \mathbb{Y})$ denotes the space of the Lipschitz continuous maps from $\mathbb{X}$ to $\mathbb{Y}$, $L$ is the loss function of the learning problem, the pair $\{x_i,y_i\}_{i=1}^{\Ntrain}$ denotes the training dataset of size $\Ntrain$, with input $x \in \mathbb{X}$ and output $y\footnote{Label $y_i \in \mathbb{R}^m$ is a vector which contains $1$ in the element that correspond to the true class and zero everywhere else.} \in \mathbb{Y}$, $\text{lip}(h)$ is the Lipschitz constant of classifier $h$, and $\alpha \in \mathbb{R}_{\geq 0}$ is the upper bound constraint on the Lipschitz constant. The classifier takes an input image of $d$ pixels and outputs a vector of probabilities of size $m$, which is the number of classes. The classifier chooses the class with the highest probability: higher probability implies higher decision confidence. We incorporate an abstain option, where the classifier abstains if the maximum probability is less than a threshold probability $p_{a}$. We consider adversarial examples, $\widetilde{x}=x+\delta$, computed as in \cite{VK-AAALM-FP:20}, where $\delta\in\mathbb{R}^d$ is a bounded perturbation ($\|\delta\|_{\infty}\leq \xi$) in the direction that induces misclassification.
%
%
\begin{figure*}[!t]
  \centering
  \includegraphics[width=\textwidth,trim={0cm 0cm 0cm
    0cm},clip]{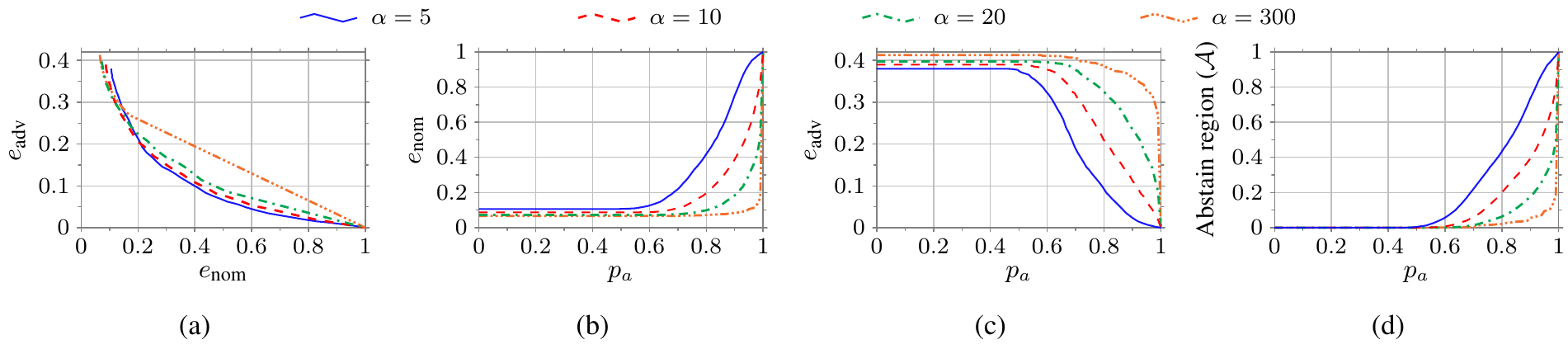}
 \caption{In the classification problem discussed in Section \ref{sec:MNIST}, $4$ classifiers are trained on the MNIST dataset using the Lipschitz-constrained loss minimization scheme in \eqref{eq:Lip_const_min}, with $\alpha=5,10,20,300$, which are represented in all $4$ panels by the solid blue line, the dashed red line, the dot-dashed green line, and the three-dot-dashed orange line, respectively. Panel (a) shows the tradeoff between $\enom$ and $\eadv$, panels (b) and (c) show $\enom$ and $\eadv$ as a function of the threshold probability, $p_a$, respectively, and panel (d) shows the ratio of the abstain region to the input space, denoted by $\mc{A}$, as a function of $p_a$. As observed in (d), the abstain region is zero for $p_a\in[0,0.5)$, it monotonically increases for $p_a\geq0.5$ till it covers the whole input space when $p_a=1$. When there is no abstaining (i.e., $p_a\in[0,0.5)$), all classifiers achieve their lowest $\enom$ and their highest $\eadv$ as observed in (b) and (c), respectively, where the classifier with $\alpha=300$ achieves the lowest $\enom$ and the highest $\eadv$ among all 4 classifiers, while the classifier with $\alpha=5$ achieves the highest $\enom$ and the lowest $\eadv$, which agrees with the tradeoff result in \cite{VK-AAALM-FP:20}. When the abstain region covers the whole input space (i.e., $p_a=1$), all classifiers achieve $\enom=1$ and $\eadv=0$ as seen in (b) and (c), respectively. Also, it is observed in (b) and (c), respectively, that as the abstain region increases (i.e., $p_a$ increases), $\enom$ increases while $\eadv$ decreases for all classifiers, which leads to the tradeoff relation between the two as observed in (a).}
%
\label{fig:tradeoff_MNIST}
\end{figure*}
\subsection{Nominal and Adversarial error}
Let $\mc{Z}=\{0,1,\dots,m-1\}$ and $\left. \widehat{\mc{Z}}=\{0,1,\dots,m-1,a\}\right.$ be the sets containing all possible true labels and all possible predicted labels by classifier $h$, respecticvely, where $a$ corresponds to the abstain option. Let $z_i\in \mc{Z}$ and $\widehat{z}_i\in \widehat{\mc{Z}}$ be the true label and the label predicted by $h$ for the input $x_i$, respectively \big(i.e., $\widehat{z}_i$ is the label that corresponds to the maximum probability in the vector $h(x_i)$, or label $a$ if the maximum probability is less than $p_a$\big). Further, let $\widetilde{z}_i\in \widehat{\mc{Z}}$ be the label predicted by $h$ for the perturbed input image $\widetilde{x}_i$. Using Definition~\ref{def: nom_error} and~\ref{def: adv_error} we compute $\enom$ and $\eadv$ for $h$ with  threshold probability $p_a$ on the testing dataset of size~$\Ntest$~as,
%
\begin{align}\label{eq:num_error}
\begin{split}
\enom(h,p_a)&=\frac{1}{\Ntest}\sum_{i=1}^{\Ntest}\1\{\widehat{z}_i\neq z_i\},\\
\eadv(h,p_a)&=\frac{1}{\Ntest}\sum_{i=1}^{\Ntest}\1\{\widetilde{z}_i\neq z_i \cap \widetilde{z}_i \neq a\},
\end{split}
\end{align}
where $\1\{\cdot\}$ denotes the indicator function.
\subsection{Nominal-Adversarial error tradeoff}
To show the implications of Theorem \ref{thrm:tradeoff}, we train four classifiers on the MNIST dataset using \eqref{eq:Lip_const_min} with $\alpha=5,10,20$, and $300$, respectively (refer to \cite{VK-AAALM-FP:20} for details about the training scheme). Then, we compute $\enom$ and $\eadv$ for each classifier using \eqref{eq:num_error} with different values of $p_a$ and a bound on the perturbation $\xi=0.3$. Fig. \ref{fig:tradeoff_MNIST} shows the numerical results on the testing dataset. Fig. \ref{fig:tradeoff_MNIST}(a) shows the tradeoff between $\enom$ and $\eadv$ for all the classifiers, which agrees with Theorem \ref{thrm:tradeoff}. Fig. \ref{fig:tradeoff_MNIST}(b)-(c) show $\enom$ and $\eadv$ as a function of $p_a$, respectively, while Fig. \ref{fig:tradeoff_MNIST}(d) shows the ratio of the abstain region to the input space, denoted by $\mc{A}$, as a function of $p_a$. As shown in Fig. \ref{fig:tradeoff_MNIST}(d), $\mc{A}$ increases at a low rate from zero to $0.1$ for $p_a\in[0.5,0.95]$ for the classifier with $\alpha=300$, then it increases at a high rate till it reaches $1$ for $p_a \in (0.95,1]$. The rate at which $\mc{A}$ increases becomes more uniform as $\alpha$ decreases, where for the classifier with $\alpha=5$, $\mc{A}$ increases with an almost uniform rate from zero at $p_a=0.5$ to $1$ at $p_a=1$. This is because as we decrease $\alpha$ in \eqref{eq:Lip_const_min}, the learned function becomes more smooth, and the change of the output probability vector over the input space becomes smoother. As observed in Fig. \ref{fig:tradeoff_MNIST}(b) and Fig. \ref{fig:tradeoff_MNIST}(c), $\enom$ increases, while $\eadv$ decreases for $p_a\in [0.5,1]$.
\section{conclusion and future work}\label{sec:conclusion}
In this work, we include an abstain option in a binary classification problem, to improve adversarial robustness. We propose metrics to quantify the nominal performance of a classifier with an abstain option and its adversarial robustness. We formally prove that, for any classifier with an abstain option, there exist a tradeoff between its nominal performance and its robustness, thus, the classifier's robustness can only be improved at the expense of its nominal performance. Further, we provide necessary conditions to design the abstain region that optimizes robustness for a desired nominal performance for $1$-dimensional binary classification problem. Finally, we~validate our theoretical results on the MNIST~dataset, where we show that the tradeoff between~performance and robustness also exist for the general~multi-class~classification~problems. This research area contains several unexplored questions including comparing tradeoffs obtained with an abstain option and tradeoffs obtained via tuning the decision boundaries, as well as investigate whether it is possible to improve the tradeoff by tuning the boundaries and the abstain region simultaneously. 

\bibliographystyle{unsrt}
\bibliography{alias,Main,FP,New}

\end{document}